\documentclass[a4paper]{article}

\usepackage{INTERSPEECH2018}
\usepackage{multirow,epstopdf,balance}

\title{Learning Acoustic Word Embeddings with Temporal Context for Query-by-Example Speech Search}
\name{Yougen Yuan$^{1}$, Cheung-Chi Leung$^{2}$, Lei Xie$^{1*}$\thanks{$^{*}$ Corresponding author}, Hongjie Chen$^{1}$, Bin Ma$^{2}$, Haizhou Li$^{3}$}
\address{
  $^1$School of Computer Science, Northwestern Polytechnical University, Xi'an, China\\
  $^2$Alibaba Inc., Singapore\\$^3$Department of Electrical and Computer Engineering, National University of Singapore, Singapore}
\email{\{ygyuan,hjchen\}@nwpu-aslp.org, lxie@nwpu.edu.cn,\\ \{cc.leung,b.ma\}@alibaba-inc.com, haizhou.li@nus.edu.sg}

\begin{document}

\maketitle
\begin{abstract}
  We propose to learn acoustic word embeddings with temporal context for query-by-example (QbE) speech search. The temporal context includes the leading and trailing word sequences of a word. We assume that there exist spoken word pairs in the training database. We pad the word pairs with their original temporal context to form fixed-length speech segment pairs. We obtain the acoustic word embeddings through a deep convolutional neural network (CNN) which is trained on the speech segment pairs with a triplet loss. By shifting a fixed-length analysis window through the search content, we obtain a running sequence of embeddings. In this way, searching for the spoken query is equivalent to the matching of acoustic word embeddings. The experiments show that our proposed acoustic word embeddings learned with temporal context are effective in QbE speech search. They outperform the state-of-the-art frame-level feature representations and reduce run-time computation since no dynamic time warping is required in QbE speech search. We also find that it is important to have sufficient speech segment pairs to train the deep CNN for effective acoustic word embeddings.
\end{abstract}
\noindent\textbf{Index Terms}: acoustic word embeddings, word pairs, temporal context, triplet loss, query-by-example spoken term detection
\section{Introduction}
\label{sec:Introduction}
Query-by-example (QbE) speech search or spoken term detection is the task of searching for the occurrence of a spoken query in search content~\cite{hazen2009query,zhang2009unsupervised}. A typical approach to this task relies on dynamic time warping (DTW) to perform acoustic pattern matching over frame-level feature representations. These feature representations can be learned in unsupervised~\cite{wang2012acoustic,yang2014intrinsic,mantena2014query,chen2016unsupervised} or supervised~\cite{tejedor2012comparison,rodriguez2014high,leung2016toward} manner. In supervised learning, classifiers are usually trained using labeled data from non-target languages to derive features.

In this paper, we propose to learn acoustic word embeddings with temporal context for QbE speech search. We assume that there exist spoken word pairs in the training database in the target language. Learning frame-level feature representations using word pairs has been shown successful~\cite{yuan2017pairwise}. However, it relies on computationally expensive DTW during the search. This prompts us to study acoustic word embeddings that encode speech at segment or word level. In this way, we can simplify the QbE speech search test by measuring the vector distance (e.\,g., cosine distance) over the acoustic word embeddings between the spoken query and the search content.

Acoustic word embeddings project speech segments into fixed-dimensional vector space where the distance between same speech content is small while the distance between different speech content is large. They have been shown successful in automatic speech recognition~\cite{bengio2014word} and isolated word discrimination~\cite{levin2013fixed,kamper2016deep,settle2016discriminative}. However, as the word boundary is not available in search content, QbE speech search is therefore considered more difficult than isolated word discrimination. To overcome this problem, studies have shown~\cite{levin2015segmental,settle2017query} that approximate nearest neighbor search over the acoustic word embeddings is a reasonable solution.

We propose the idea to include the leading and trailing word sequences as the temporal context of a word. The word pairs are padded with their original temporal context to form fixed-length speech segment pairs. We train a deep convolutional neural network (CNN) with a triplet loss using the speech segment pairs to learn acoustic word embeddings. During QbE speech search, we propose to shift a fixed-length analysis window to obtain a sequence of embeddings on search content, then we search over the embeddings instead of frame-level feature representations to find the matching spoken query.

The novel contribution of this paper is that, for the first time, we incorporate the temporal context to improve the acoustic word embeddings for QbE speech search. With the temporal context, we learn the possible neighboring speech sequences around the words, which reduces the mismatch between the learning of embeddings and its application on the search content. Our proposed technique outperforms the state-of-the-art frame-level feature representations~\cite{yuan2017pairwise} and reduces the run-time computation since no DTW is required in QbE speech search. As learning acoustic word embeddings requires a larger number of word pairs than learning frame-level feature representations~\cite{yuan2016learning}, we use more word pairs discovered from the Switchboard speech corpus and we test the effect of number of speech segment pairs in learning acoustic word embeddings for QbE speech search.
\vspace{-2.65mm}
\section{QbE speech search using acoustic word embeddings}
\label{sec:QbE speech search using acoustic word embeddings}
\subsection{Learning embeddings with temporal context}
\label{subsec:Learning embeddings with temporal context}
Acoustic word embeddings which are extracted from a typical feed-forward deep neural network usually require the input with fixed-length. Zero padding has been shown successful in learning acoustic word embeddings for isolated word discrimination~\cite{kamper2016deep,yuan2016learning}. With zero padding, all speech segments are padded with zeros on both side of each segment to the same length. In the case of QbE speech search, because the word boundary is not available, it is hard to segment the search content into isolated words. Therefore, we propose to shift a fixed-length analysis window to segment the search content into many fixed-length speech segments. As these speech segments may contain sub-words, one or more whole words, there exists a mismatch between the learning of embeddings and the use of embeddings at run-time. To mitigate such mismatch, we propose to use the temporal context of a word to learn acoustic word embeddings in QbE speech search.

The temporal context refers to the original leading and trailing word sequences on both sides of a word. The way we incorporate the temporal context is also called temporal context padding. As illustrated in Fig. 1, given a word instance ${x_a}^{w}$, we add its original previous word sequence as ${x_a}^{w+}$ in front and its subsequent word sequence as ${x_a}^{w-}$ behind with the same number of frames. Notice that the temporal context may contain a partial word (e.\,g., ``wi-'' in ``with''), a whole word (e.\,g., ``here''), or multiple words (e.\,g., ``a real''). We assume that word pairs (e.\,g., $({x_a}^w, {x_p}^w)$) identified by humans are available. The word pairs are padded with their original temporal context to form speech segment pairs (e.\,g., $(x_a, x_p)$). In this way, the speech segment $x_a$ contains the same word (e.\,g., ``problem'') as the speech segment $x_p$, while the speech segment $x_n$ contains a different word (e.\,g., ``nowadays'').
\begin{figure}[t]
	\centering
	\includegraphics[width=\linewidth]{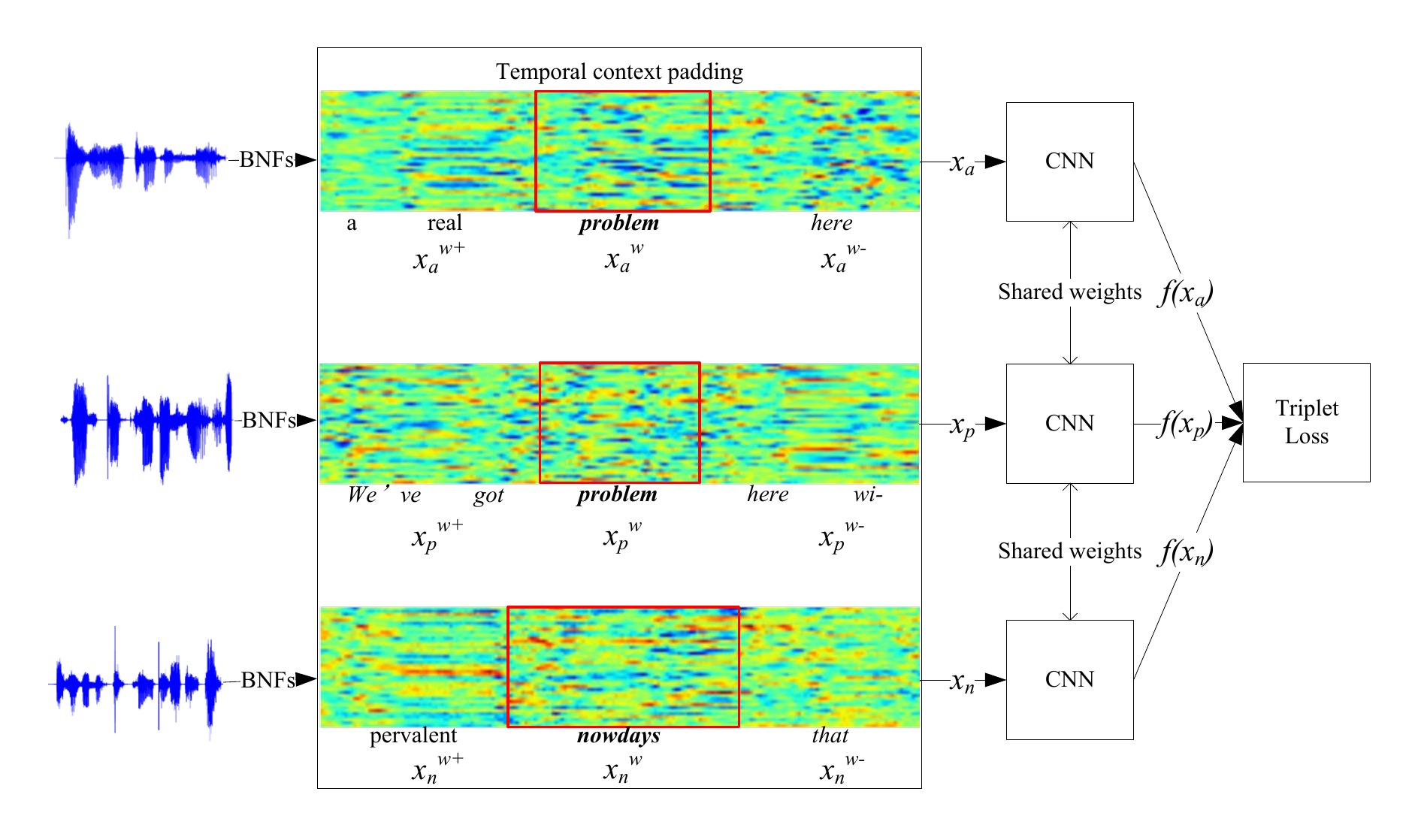}
	\caption{The diagram of learning acoustic word embeddings with temporal context padding.}
	\label{fig:Triplet network}
\end{figure}

We employ deep neural networks with a triplet loss to learn acoustic word embeddings. The deep neural networks take the triplets as input. Each triplet consists of 3 examples $({x_p},{x_a},{x_n})$. We use a pair of speech segments as an anchor example $x_a$ and a positive example $x_p$, and we randomly sample another speech segment as a negative example $x_n$. Learning acoustic word embeddings is shown in Fig~\ref{fig:Triplet network}. Each example is represented by multi-lingual bottleneck features (BNFs), which capture rich information of phonetic discrimination from other language resources. We aim to learn a function $f$ which maps an example $x$ to the fixed-dimensional embedding $f(x)$. The deep CNN, which has been shown successful to learn this function in isolated word discrimination~\cite{kamper2016deep,yuan2016learning}, is used here for learning embeddings with temporal context.

The triplet loss was originally proposed in~\cite{schroff2015facenet} for learning face embeddings from image. As for learning acoustic word embeddings from speech, we aim to increase the similarity between the embeddings $(f({x_p}),f({x_a}))$, while decreasing the similarity between the embeddings $(f({x_n}),f({x_a}))$. Our triplet loss is defined as 
\begin{equation}
Loss({x_p},{x_a},{x_n})=max\{0,\delta + d^+ - d^-\}	
\label{Loss}
\end{equation}
\begin{equation}
d^+=\frac{1- \frac{f({x_p})\cdot f({x_a})}{{\Arrowvert f({x_p}) \Arrowvert}_2{\Arrowvert f({x_a}) \Arrowvert}_2 }}{2}
\label{d-}
\end{equation}
\begin{equation}
d^-=\frac{1- \frac{f({x_n})\cdot f({x_a})}{{\Arrowvert f({x_n}) \Arrowvert}_2{\Arrowvert f({x_a}) \Arrowvert}_2 }}{2}
\label{d+}
\end{equation}
where $\delta$ is a margin constraint that regularizes the gap between the cosine distance of same-word embeddings $d^+$ and the cosine distance of different-word embeddings $d^-$. We set the margin to 0.15 as in~\cite{kamper2016deep,yuan2016learning}. The acoustic word embeddings $f(x)$ are extracted from the last layer of the trained deep CNN.
\begin{figure}[t]
	\centering
	\includegraphics[width=\linewidth]{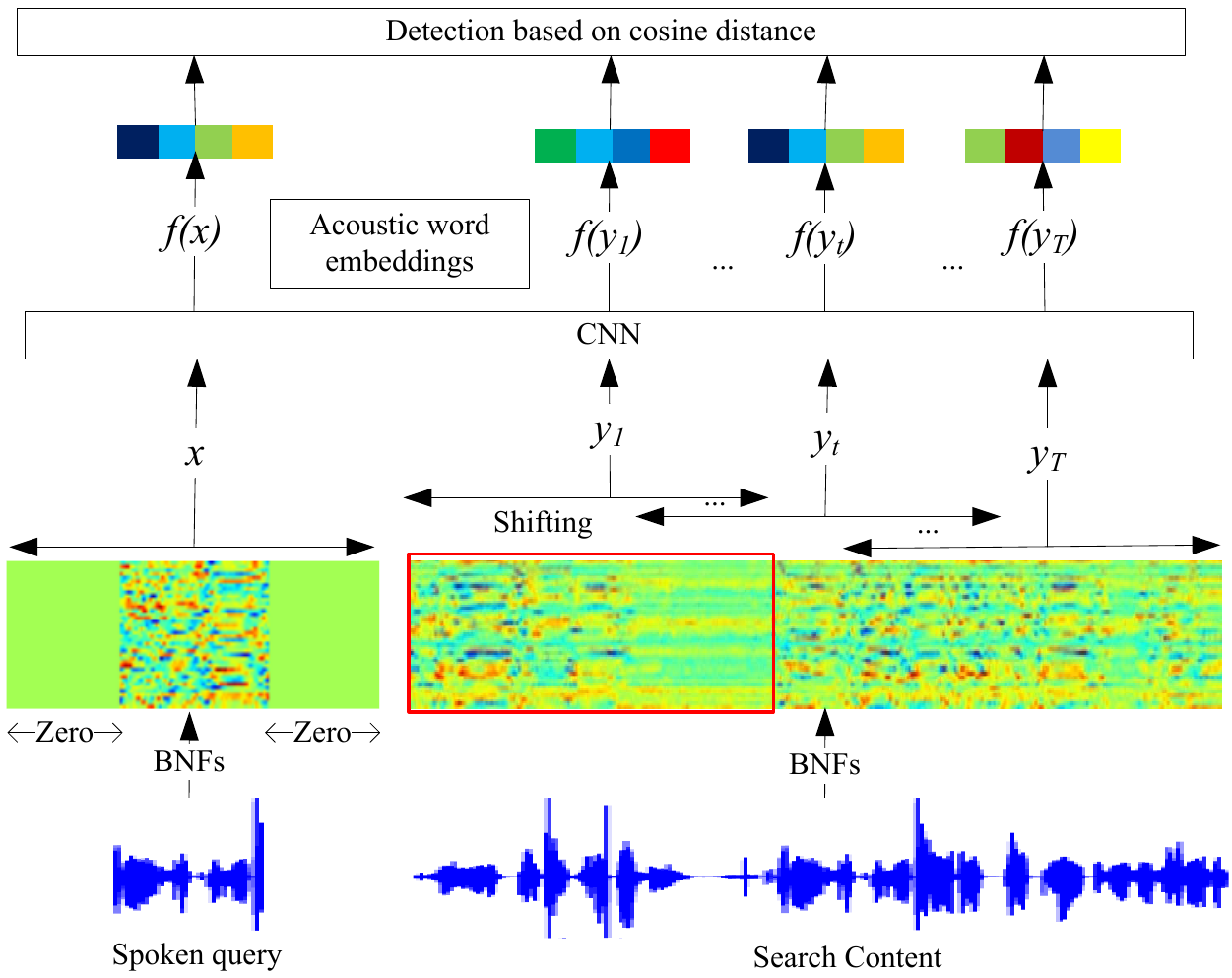}
	\caption{QbE speech search using acoustic word embeddings.}
	\label{fig:QbE speech search}
\end{figure}
\begin{figure*}[t]
	\centering
	\includegraphics[width=\linewidth]{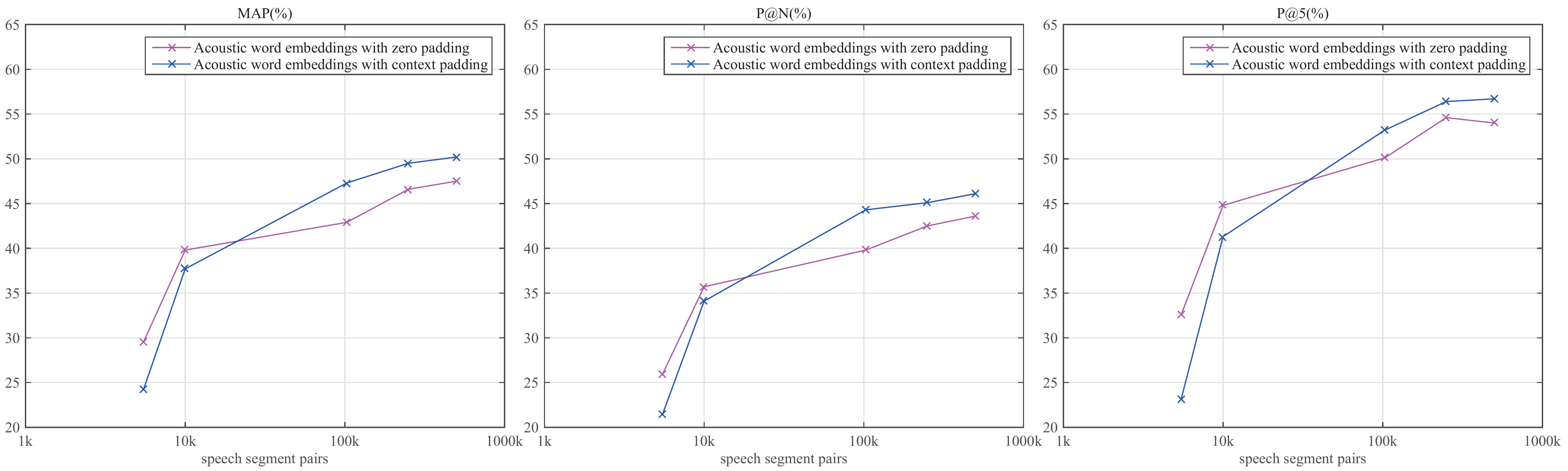}
	\caption{Comparative study between zero padding and context padding in acoustic word embeddings for QbE speech search. Multi-lingual BNFs and Set 2 are used.}
	\label{fig:padding}
\end{figure*}
\subsection{Shifting analysis window on search context}
\label{subsec:QbE speech search}
Fig.~\ref{fig:QbE speech search} illustrates the process of our proposed QbE speech search system using acoustic word embeddings. As an indexing process, we propose to apply a fixed-length analysis window on the search content $y$ by shifting along the time axis. The speech segment in the analysis window is then converted into an acoustic word embedding by the trained deep CNN. As a result, the search content is indexed by a sequence of acoustic word embeddings as $(f(y_1),...,f(y_i),...,f(y_T))$. As no context information is available for the spoken query $x$, we pad zeros to both sides of $x$ as the input to the trained deep CNN to obtain the embedding $f(x)$. In this way, the vector distance, instead of DTW distance, can be directly used over the embeddings between the spoken query and the search content. 

Note that the mismatch between the embeddings in the spoken query and the search content still exist. To further mitigate the mismatch, we also considered different ways to learn the embeddings, including zero padding, and combining context padding and zero padding (using the two padding methods in a speech segment triplet or using different padding methods in different triplets). However these ways do not further improve the search accuracy in our preliminary test.

The size of the fixed-length analysis window is determined by the average length of all speech segments used in learning acoustic word embeddings. We set the window shift size to 5 frames as we find that a shifting smaller than 5 frames doesn't improve. A minimum distance cost can be calculated by:
\begin{equation}
Cost(x,y)=min(1- \frac{f(x)\cdot f({y_i})}{{\Arrowvert f(x) \Arrowvert}_2{\Arrowvert f({y_i}) \Arrowvert}_2 }), i=1,...,T
\label{eq:cost}
\end{equation}
Given a spoken query, all the minimum distance costs in search content are returned by the QbE speech search system.
\section{Experiments}
\label{sec:Experiments}
\subsection{Experimental setup}
\label{subsec:Experimental setup}		
\begin{table*}[!t]
	\caption{Comparison of different feature representations for QbE speech search. 500$k$ speech segment pairs in Set 2 are used.}
	\label{tab:QbE speech search}
	\centering
	\begin{tabular}{lcccccc}
		\toprule
		\multirow{2}*{Representation} & Input features   &  Use & Run-time computation & MAP & P@N  & P@5  \\
		& of paired examples  &  DTW? & (seconds) &  &   &   \\
		\midrule
		Multi-lingual BNFs & N/A  & Yes & 4,752 & 0.400 & 0.365 & 0.485 \\				
		Frame-level feature representations~\cite{yuan2017pairwise}  & Multi-lingual BNFs  & Yes & 9,506 & 0.485  & 0.446 & 0.566 \\	
		Acoustic word embeddings (proposed) & Multi-lingual BNFs  & No & \textbf{1,017} & \textbf{0.502}  & \textbf{0.462} & \textbf{0.567} \\
		\bottomrule
	\end{tabular}	
\end{table*}	
To evaluate the effectiveness of our proposed acoustic word embeddings, we conducted the QbE speech search on the English Switchboard corpus. From our previous work~\cite{yuan2016learning}, we observed that learning word-level embeddings should require a larger number of word pairs than learning frame-level feature representations. Thus we extended two training sets: 
\begin{itemize}
	\item Set 1: It has the same vocabulary size (1,687) as in~\cite{yuan2016learning}, but it consists of 37$k$ word instances (about 6.6 hours of speech) that make up 500$k$ speech segment pairs with temporal context padding.
	\item Set 2: The vocabulary size is increased to 5,476, and the dataset consists of 53$k$ word instances (about 9.5 hours of speech) that also make up 500$k$ speech segment pairs with temporal context padding.
\end{itemize}
We used the same development set as in~\cite{jansen2013weak,kamper2015unsupervised,kamper2016deep,yuan2016learning,yuan2017pairwise} for learning acoustic word embeddings. As for QbE speech search tests, we followed the data setting in~\cite{yuan2017pairwise}. We used 346 spoken queries as the keyword set and 100 utterances as the test set.

We included the leading and trailing word sequences as the temporal context of a word to form a speech segment with the length of 200 frames (2 seconds). All the speech segments were represented by 40-dimensional multi-lingual BNFs as in~\cite{yuan2017pairwise}. The BNF extractor was trained using Mandarin Chinese and Spanish telephone speech. We trained a deep CNN with a triplet loss using speech segment pairs to learn acoustic word embeddings. The deep CNN model consists of two convolutional and max pooling layers, a fully-connected ReLU layer with 2,048 hidden units and a fully-connected linear layer with 1,024 hidden units. We implemented the model using the Theano toolkit~\cite{bergstra2010theano}, and we trained the model using stochastic gradient descent with the mini-batch size of 1,024. All the neural weights were initialized randomly. An ADADELTA~\cite{zeiler2012adadelta} optimizer was used with the momentum of $\rho=0.9$ and the precision of $\epsilon=10^{-6}$. Training would be terminated if the loss on the development set was not improved over 20 epochs. 

As in~\cite{wang2012acoustic,chen2016unsupervised,yuan2017pairwise}, three different evaluation metrics are used for QbE speech search: 1) mean average precision (MAP), which is the mean of average precision for each query on search content. 2) Precision of the top N utterances in the test set (P@N), where N is the number of target utterances involving the query term. 3) Precision of the top 5 utterances in the test set (P@5).	
\subsection{Temporal context in acoustic word embeddings}
\label{subsec:Temporal padding in acoustic word embeddings}
To validate the efficiency of our proposed acoustic word embeddings learned with temporal context  for QbE speech search, we compared context padding with zero padding in learning acoustic word embeddings based on multi-lingual BNFs. From Fig.~\ref{fig:padding} we can find that context padding outperforms zero padding when more than 10$k$ speech segment pairs are available in Set 2 (about 3\%-11\% relative improvement in three evaluation metrics). Similar results are also obtained in Set 1 with a smaller vocabulary. The experiment results suggest that the temporal context padding learns the possible neighboring speech sequences around the words, which reduces the mismatch between the learning of embeddings and the use of embeddings on the search content for QbE speech search.
\begin{figure*}[!t]
	\centering
	\includegraphics[width=\linewidth]{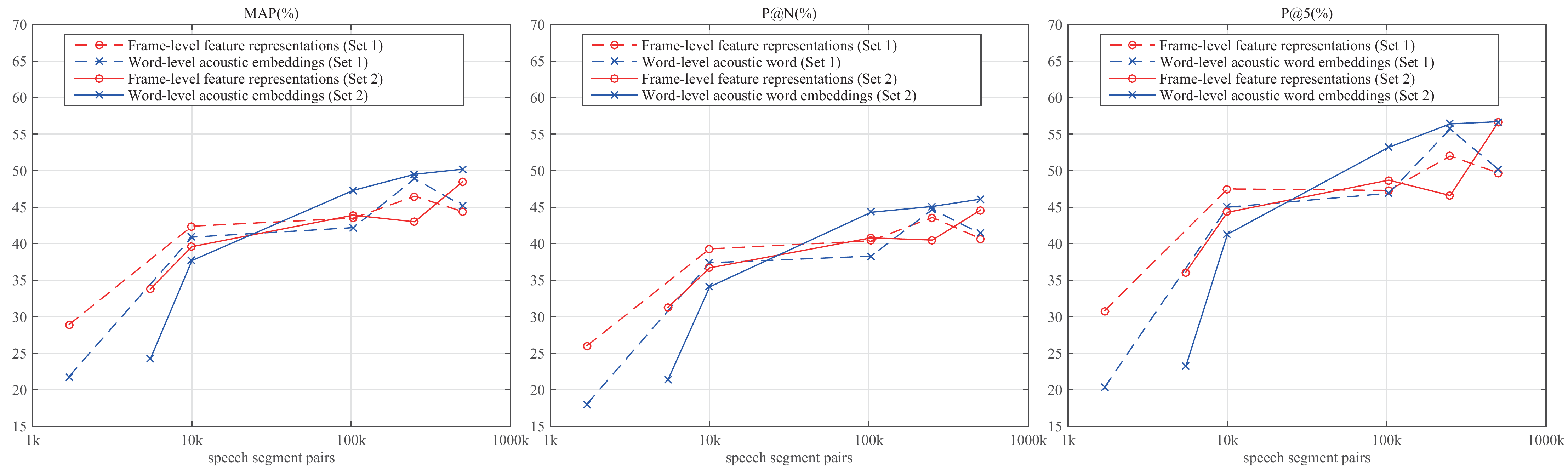}
	\caption{Effect of number of speech segment pairs in learning speech representations for QbE speech search.}
	\label{fig:speech segment pairs}
\end{figure*}
\subsection{Comparison of different feature representations}
\label{subsec:Experimental results}
Table~\ref{tab:QbE speech search} lists the performance of QbE speech search using different feature representations, including multi-lingual BNFs, the state-of-the-art frame-level feature representations in~\cite{yuan2017pairwise} and our proposed acoustic word embeddings learned with temporal context. These feature representations are trained using 500$k$ speech segment pairs in Set 2. The QbE speech search based on these feature representations are tested using a single thread on a workstation equipped with an Intel Xeon E5-2680 @ 2.7GHz CPU. The results show that the acoustic word embeddings outperform the frame-level feature representations (about 4\% relative improvement in both MAP and P@N), and they reduce run-time computation since no dynamic time warping is required in QbE speech search. This suggests that learning acoustic word embeddings with temporal context is effective in terms of both accuracy and computational efficiency. In addition, we also find that the acoustic word embeddings based on multi-lingual BNFs outperform those based on spectral features (e.\,g., mel-frequency cepstral coefficients). This demonstrates that multilingual knowledge from resource-rich languages is helpful to learn acoustic word embeddings for QbE speech search.						
\subsection{Effect of number of speech segment pairs}
\label{subsec:Effect of number of speech segment pairs}
We also investigated how the number of speech segment pairs in learning speech representations would affect the performance for QbE speech search. We randomly selected subsets of $N$=[$M$, 10$k$, 100$k$, 250$k$, 500$k$] speech segment pairs, where $M$ represents the minimum speech segment pairs in Set 1 and Set 2 respectively. The evaluation results are plotted in Fig.~\ref{fig:speech segment pairs}. 

We observe that acoustic word embeddings derived from Set 1 and Set 2 consistently improve the search results as the number of speech segment pairs increases. When we have more than 10$k$ speech segment pairs, the acoustic word embeddings of Set 2 consistently outperform those of Set 1. This suggests that we can train a better deep CNN for acoustic word embeddings using a larger vocabulary, and it is important to have sufficient speech segment pairs to learn acoustic word embeddings for QbE speech search. In addition, we also reported the results of frame-level feature representations trained in both datasets. From Fig.~\ref{fig:speech segment pairs} we can find that our proposed acoustic word embeddings can consistently give higher search accuracies than the frame-level feature representations when more than 100$k$ speech segment pairs are available. 
\section{Conclusion}
\label{sec:Conclusion}
We have proposed a novel approach to learn convolutional neural acoustic word embeddings trained with temporal context padding for QbE speech search. The temporal context padding reduces the mismatch between the learning of embeddings and the use of embeddings on search content. Our proposed acoustic word embeddings can outperform the state-of-the-art frame-level feature representations and reduce run-time computation since no dynamic time warping is required in QbE speech search. Sufficient speech segment pairs with sufficient vocabulary coverage are important to learn acoustic word embeddings for QbE speech search. In the future, the discovery and selection of speech segment pairs will be worth exploring, and we will investigate recurrent neural networks, which are capable of modeling sequences, to obtain better acoustic word embeddings for this task.
\section{Acknowledgments}
\label{Acknowledgments}
This work was supported by the National Natural Science Foundation of China (Grant No. 61571363) and the China Scholarship Council (Grant No. 201706290169).
\newpage
\balance
\bibliographystyle{IEEEtran}
\bibliography{mybib}
\end{document}